\documentclass[11pt,a4paper]{article}
\usepackage{emnlp2018}
\usepackage{times}
\usepackage{latexsym}

\usepackage{url}
\usepackage{hyperref}

\aclfinalcopy 

\usepackage{bm} 
\usepackage{amsopn} 
\usepackage{mathtools} 
\usepackage{amssymb} 
\usepackage{amsmath}
\usepackage{amsfonts}
\usepackage{isomath}
\usepackage{dsfont}

\allowdisplaybreaks 

\newcommand{\ie}{i.\,e., }

\newcommand{\vs}{vs. }

\newcommand{\class}{y}

\newcommand{\example}{\bm{x}}

\newcommand{\weights}{\bm{\theta}}
\DeclareMathOperator*{\argmax}{arg\,max}
\DeclareMathOperator*{\argmin}{\arg\!\min}

\usepackage{algorithm}
\usepackage{algorithmic}

\newcommand{\algorithmicbreak}{\textbf{break}}
\newcommand{\BREAK}{\STATE \algorithmicbreak}

\usepackage{multirow}
\usepackage{siunitx} 
\usepackage{etoolbox} 
\robustify\bfseries
\usepackage{booktabs}

\newcommand \eye {\matrixsym{I}}

\newcommand \MX {\matrixsym{X}}

\newcommand \vx {\bm{x}}
\newcommand \vy {\bm{y}}

\newcommand \vf {\bm{f}}

\newcommand \vr {\bm{r}}

\newcommand\Reals {{\mathds{R}}}
 
\usepackage{bbm}

\newcommand \defn {\mathrel{\triangleq}}
\newcommand\Set {{\mathcal{I}}}

\newcommand{\pluseq}{\mathrel{+}=}

\newcommand{\superscript}[1]{\ensuremath{^{\textrm{#1}}}}

\usepackage{tikz}
\usetikzlibrary{shapes,arrows}

\usetikzlibrary{calc,matrix,positioning}
\usetikzlibrary{decorations.pathmorphing}
\usepackage{amsthm}
\usepackage{colortbl}

\usepackage[first=0,last=9]{lcg}

\definecolor{LightCyan}{rgb}{0.88,1,1}
\definecolor{applegreen}{rgb}{0.55, 0.71, 0.0}
\definecolor{azure}{rgb}{0.0, 0.5, 1.0}
\definecolor{darkpastelgreen}{rgb}{0.01, 0.75, 0.24}
\definecolor{brightube}{rgb}{0.82, 0.62, 0.91}
\definecolor{coralred}{rgb}{1.0, 0.25, 0.25}

 \definecolor{capri}{RGB}{97, 177, 255}
\definecolor{myred}{RGB}{255, 76, 76}
\definecolor{mygreen}{RGB}{52, 168, 83}
\definecolor{myblue}{RGB}{66, 133, 244}
\definecolor{myorange}{RGB}{255, 165, 0}
\definecolor{myyellow}{RGB}{251,188,5}

\usepackage{balance}

\usetikzlibrary{arrows.meta,shapes.arrows,chains,decorations.pathreplacing}

\tikzset{%
  MyStyle/.style={draw, text width=35pt, text height=30pt, text centered,minimum height=1mm)},
  myarrow/.style={shape=single arrow, rotate=90, inner sep=5pt, outer sep=0pt, single arrow head extend=0pt, minimum height=7.5pt, text width=4pt, draw=blue!50, fill=blue!25}
}

\newcommand{\blockmatrix}[9]{
  \draw[draw=#4,fill=#5] (0,0) rectangle( #1,#2);
  \ifthenelse{\equal{#6}{true}}
  {
    \draw[draw=#7,fill=#8] (0,#2) -- (#9,#2) -- ( #1,#9) -- ( #1,0) -- ( #1 - #9,0) -- (0,#2 -#9) -- cycle;
  }
  {}
  \draw ( #1/2, #2/2) node { #3};
}

\newcommand{\mblockmatrix}[4][none]{
  \begin{tikzpicture} 
  \ifthenelse{\equal{#1}{none}}
  {
    \blockmatrix{#2}{#3}{#4}{none}{none}{false}{none}{none}{0.0}
  }
  {
    \definecolor{fillcolor}{rgb}{#1}
    \blockmatrix{#2}{#3}{#4}{none}{fillcolor}{false}{none}{none}{0.0}
  }
  \end{tikzpicture}
}

\newcommand{\fblockmatrix}[4][none]{
  \begin{tikzpicture} 
  \ifthenelse{\equal{#1}{none}}
  {
    \blockmatrix{#2}{#3}{#4}{black}{none}{false}{none}{none}{0.0}
  }
  {
    \definecolor{fillcolor}{rgb}{#1}
    \blockmatrix{#2}{#3}{#4}{black}{fillcolor}{false}{none}{none}{0.0}
  }
  \end{tikzpicture}
}

\newcommand{\dblockmatrix}[4][none]{
  \begin{tikzpicture} 
  \ifthenelse{\equal{#1}{none}}
  {
    \blockmatrix{#2}{#3}{#4}{black}{none}{true}{black}{none}{0.35cm}
  }
  {
    \definecolor{fillcolor}{rgb}{#1}
    \blockmatrix{#2}{#3}{#4}{black}{none}{true}{black}{fillcolor}{0.35cm}
  }
  \end{tikzpicture}
}

\newcommand{\diagonalblockmatrix}[5][none]{
  \begin{tikzpicture} 

  \ifthenelse{\equal{#1}{none}}
  {
    \blockmatrix{#2}{#3}{#4}{black}{none}{true}{black}{none}{#5}
  }
  {
    \definecolor{fillcolor}{rgb}{#1}
    \blockmatrix{#2}{#3}{#4}{black}{none}{true}{black}{fillcolor}{#5}
  }

  \end{tikzpicture}
}

\usetikzlibrary{er,positioning}

\title{Orthogonal Matching Pursuit for Text Classification}

\author{Konstantinos Skianis\textsuperscript{1}, Nikolaos Tziortziotis\textsuperscript{1}, Michalis Vazirgiannis\textsuperscript{1,2} \\
\textsuperscript{1}LIX, \'Ecole Polytechnique, France \\
\textsuperscript{2}Athens University of Economics and Business, Greece \\
  {\tt \{kskianis, mvazirg\}@lix.polytechnique.fr, ntziorzi@gmail.com}} 
\date{}

\begin{document}
\maketitle

\begin{abstract}
In text classification, the problem of overfitting arises due to the high dimensionality, making regularization essential.
Although classic regularizers provide sparsity, they fail to return highly accurate models.
On the contrary, state-of-the-art group-lasso regularizers provide better results at the expense of low sparsity.
In this paper, we apply a greedy variable selection algorithm, called Orthogonal Matching Pursuit, for the text classification task.
We also extend standard group OMP by introducing overlapping Group OMP to handle overlapping groups of features.
Empirical analysis verifies that both OMP and overlapping GOMP constitute powerful regularizers, able to produce effective and very sparse models.
Code and data are available online\footnote{\href{https://github.com/y3nk0/OMP-for-Text-Classification}{github.com/y3nk0/OMP-for-Text-Classification}}.
\end{abstract}

\section{Introduction} \label{introduction}

The overall high dimensionality of textual data is of major importance in text classification (also known as text categorization), opinion mining, noisy text normalization and other NLP tasks.
Since in most cases a high number of words occurs, one can easily fall in the case of overfitting.
Regularization remains a key element for addressing overfitting in tasks like text classification, domain adaptation and neural machine translation \cite{DBLP:journals/taslp/ChenR00,lu2016general,barone2017regularization}.
Along with better generalization capabilities, a proper scheme of regularization can also introduce sparsity.
Recently, a number of text regularization techniques have been proposed in the context of deep learning \cite{qian2016linguistically,ma2017group,zhang2017exploiting}.

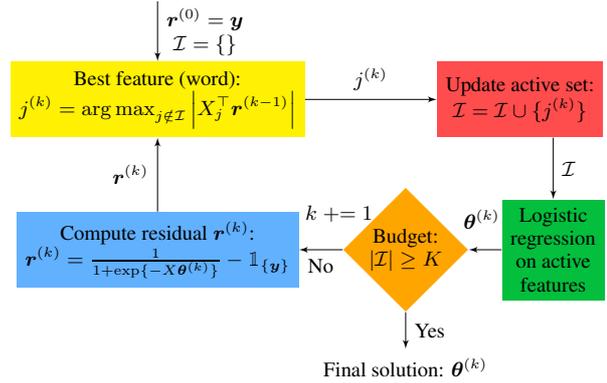
\begin{figure}[t]
\centering
    \resizebox{\columnwidth}{!}{%
\hspace{-0.8cm}
\tikzstyle{block} = [fill=blue!20, rectangle, 
    minimum height=3em, minimum width=2em]
\tikzstyle{sum} = [node distance=1cm]
\tikzstyle{input} = [coordinate]
\tikzstyle{output} = [coordinate]
\tikzstyle{pinstyle} = [pin edge={to-,thin,black}]

\begin{tikzpicture}[auto, node distance=2.3cm,>=latex',font=\small]
    \node [input, name=input] {};
    
    \node [block, below of=input,fill=yellow,node distance=1.5cm,align=center] (best) {
    Best feature (word):\\
    $j^{(k)} = \argmax_{j \notin \Set} \left| \MX_j^{\top} \vr^{(k-1)}\right|$
    };
    \node [block, right of=best, fill=myred,node distance=5.5cm,align=center] (system) {
    
  Update active set:
   \\ $\Set = \Set \cup \{j^{(k)}\}$
    };
    \draw [->] (input) -- node[name=r,align=center] {$\vr^{(0)} = \vy$\\ $\Set=\{\}$} (best);
    \draw [->] (best) -- node[name=u] {$j^{(k)}$} (system);
    \node [block, draw=none, fill=darkpastelgreen, below of=system, left=-2.55cm of system, align=center] (l2) {Logistic\\ regression\\on active\\features};
    \node [relationship, draw=none, fill=myorange, left = 0.5cm of l2,node distance=2cm,align=center,name=check] (check) {Budget:\\$|\Set|\geq K$};
    \node [block, draw=none, fill=capri, below of=best,align=center] (residual) {Compute residual $\vr^{(k)}$: \\
    	$\vr^{(k)} =  \frac{1}{1 + \exp\{-\MX \weights^{(k)}\}} - \mathbbm{1}_{\{\vy\}}$
    };
    \node [below of=check, node distance=1.8cm,name=output] {Final solution: $\weights^{(k)}$};
	
    
    \draw [->] ($ (system.south east)!0.6!(system.south) $) -- node[align=center]{$\Set$} (l2);
    \draw [->] (l2) -- node[name=l2link,align=center] {} (check);
    \draw [->] (check) -- node[name=no,align=center]{No} (residual);
    \node[above of=no, right=-0.7cm of no, node distance=0.8cm] {$k \pluseq 1$};
    \node[above of=l2link, right=-0.6cm of l2link, node distance=0.6cm] {$\weights^{(k)}$};
     \draw [->] (residual) -- node[name=r]{$\vr^{(k)}$} (best);
     \draw [->] (check) -- node[name=out,align=center]{Yes} (output);

\end{tikzpicture}
}
\vspace{-0.8cm}
\caption{OMP pipeline where $\MX \in \Reals^{N\times d}$ is the design matrix, $\vy \in \Reals^{N}$ is the response vector, $K$ is our budget and $\Set$ the set of active features.}
\label{fig:pipeline}
\end{figure}

Apart from $\ell_1$, $\ell_2$ and elastic net, a very popular method for regularizing text classification is group lasso. 
\citet{yogatama_making_2014} introduced a group lasso variant to utilize groups of words for logistic models.
Occasionally though, these groupings are either not available or hard to be extracted.
Moreover, no ground truth groups of words exist to validate their quality.
Furthermore, group lasso can also fail to create sparse models.
Lastly, there has been little work in overlapping group regularization for text, since words can appear in different groups, following the intuition that they can share multiple contexts or topics.

In this work, we apply two greedy variable selection techniques to the text classification task, Orthogonal Matching Pursuit (OMP) and overlapping group Orthogonal Matching Pursuit (GOMP).
In the case of GOMP, we build upon work of \citet{LozanoSA11}, where the authors propose the GOMP algorithm for Logistic Regression for selecting relevant groups of features.
More specifically, standard GOMP is based on the assumption that a number of disjoint groups of features are available.
Nevertheless, in most cases, these groups are not disjoint.
To overcome this problem we extend GOMP to handle overlapping groups of features. 
We empirically show that both OMP and overlapping GOMP provide highly accurate models, while producing very sparse models compared to group lasso variants.
Figure \ref{fig:pipeline} illustrates schematically the pipeline of OMP. 

Our contribution can be summarized in the following novel aspects: 
(1) apply OMP to text classification;
(2) introduce overlapping GOMP, moving from disjoint to overlapping groups;
(3) analyze their efficiency in accuracy and sparsity, compared to group lasso variants and state-of-the-art deep learning models.

The rest of the paper is organized as follows.
Section~\ref{sec:background} presents the background about the classification task, and Section~\ref{related} gives an overview of the related work.
Section~\ref{sec:OMP} formally introduces the proposed OMP and overlapping GOMP algorithms for the text classification problem.
Experimental results are presented in Section~\ref{experiments}. 
We conclude the paper in Section~\ref{conclusion} by discussing possible future directions.

\section{Background \& Notation}
\label{sec:background}

In this section, we set the theoretical and practical background, needed to tackle text classification.

\subsection{Loss minimization}

In the binary classification problem, the objective is to assign an instance vector $\vx \in \Reals^d$, which represents a document in our setting, to a binary response $y \in \{-1,1\}$. In text classification, $d$ represents the size of our dictionary concatenated with an additional bias term.  

We begin by transforming the classification problem into a loss minimization problem.
For that purpose, a loss function should be defined that quantifies the loss between the prediction of a classifier and the true class label, $y_i$, associated with a specific document (instance), $\vx_i$.

Logistic regression models the class conditional probability, as:
\begin{equation}
   P(Y = y| \vx) = \frac{1}{1+\exp\{-y (\weights^{\top} \vx)\}},
\end{equation}
where vector $\weights$ contains the unknown model's parameters, and the hyperplane $\weights^{\top} \vx = 0$ is the decision boundary of the classifier that separates the two classes. Given a training set of i.i.d. data point $\{(\vx_i, y_i)\}_{i=1}^{N}$, we find the optimal model's parameters, $\weights^*$, by minimizing the negative log likelihood:
\begin{equation}
  \weights^* = \argmin_{\weights} \sum_{i=1}^{N}{\mathcal{L}(\example_i, \weights, y_i)},
\end{equation}
where $\mathcal{L}(\example_i, \weights, y_i) = \log[1+\exp\{-y_i (\weights^{\top} \example_i)\}]$
is the loss function of our model. 
It should be also noticed that other loss functions can be used such as hinge loss, square loss, etc.. 
For linear classifiers such as Linear Least Square Fit, Logistic Regression and linear Support Vector Machines, in the case of binary predictions the $\mathcal{L}(\example, \weights, \class)$ is respectively $[1-\class (\weights^\top\example)]^2$ (squared loss), $\log[1+\exp\{-y (\weights^{\top} \example)\}$ (log loss) and $[1 - \class (\weights^{\top} \vx)]_+$ (hinge loss).

\subsection{Regularization}

By only minimizing the empirical risk, a model can be led to severe overfitting in the case where the number of the features (dictionary) is much higher than the number of the instances (documents) in training set. In practice, it yields models with poor generalization capabilities (i.e., lower performances on the test set) that fit the noise contained in the training dataset instead of learning the underlying pattern we are trying to capture. 
Additionally, if two hypothesis lead to similar low empirical risks, one should select the ``simpler'' model for better generalization capabilities.
\paragraph{Interpretation}
The concept of regularization encompasses all these ideas.
It can be interpreted as taking into account the model complexity by discouraging feature weights from reaching large values; incorporating prior knowledge to help the learning by making prior assumptions on the feature weights and their distribution; and helping compensate ill-posed conditions.

\paragraph{Expected risk}
Regularization takes the form of additional constraints to the minimization problem, \ie a budget on the feature weights, which are often relaxed into a penalty term $\Omega(\weights)$ controlled via a Lagrange multiplier $\lambda$ (see \citet{boyd_convex_2004} for more details about the theory behind convex optimization).
Therefore, the overall expected risk \citep{vapnik_principles_1991} can be expressed as the weighted sum of two components: the empirical risk and a penalty term,  called ``Loss+Penalty" \cite{hastie_elements_2009}.
In this way, the optimal set of feature weights $\weights^*$ is found as:
\begin{equation}
  \weights^* = \argmin_{\weights} \sum_{i=1}^{N}{\mathcal{L}(\example_i, \weights, y_i)} + \vphantom{\sum_{i=1}^{N}{}}\lambda\Omega(\weights)
\end{equation}
where the free parameter $\lambda \geq 0$ governs the importance of the penalty term compared with the loss term.

\section{Related Work}  \label{related}

In this section, we review particularly relevant prior work on regularization for text classification and more specifically methods based on group lasso.

In many applications of statistics and machine learning, the number of exploratory variables may be very large, while only a small subset may truly be relevant in explaining the response to be modelled. 
In certain cases, the dimensionality of the predictor space may also exceed the number of examples. 
Then the only way to avoid overfitting is via some form of “capacity control” over the family of dependencies being explored. 
Estimation of sparse models that are supported on a small set of input variables is thus highly desirable, with the additional benefit of leading to parsimonious models, which can be used not only for predictive purposes but also to understand the effects (or lack thereof) of the candidate predictors on the response.

More specifically, regularization in text scenarios is essential as it can lead to removing unnecessary words along with their weights.
For example, in text classification, we may only care for a small subset of the vocabulary that is important during the learning process, by penalizing independently or in grouped way noisy and irrelevant words.

With noiseness we refer to user-generated words that may increase the dimensionality and complexity of a problem, while having a clear decreasing effect in performance.

Another example task is text normalization, where we want to transform lexical variants of words to their canonical forms. 
Text normalization can be seen as a machine learning problem \cite{ikeda2016japanese} and thus regularization techniques can be applied.

Next we present standard regularization methods, which prove to be effective for classification tasks.
We also use them later as baselines for our experiments.

\paragraph{$\ell_1$, $\ell_2$ regularization}
Two of the most used regularization schemes are $\ell_1$-regularization, called \textit{Lasso} \cite{tibshirani_regression_1996} or \textit{basis pursuit} in signal processing \cite{chen2001atomic}, and $\ell_2$-regularization, called \textit{ridge} \cite{hoerl_ridge_1970} or \textit{Tikhonov} \cite{tikhonov_solutions_1977}, which involve adding a penalty term ($\ell_1$ and $\ell_2$ norms of the parameter vector, respectively) to the error function: \vspace{-1em}
\begin{align}
  \Omega_{lasso}(\weights) &= \sum_{i=1}^d |\theta_i| = \| \weights \|_1,\\
  \Omega_{rigde}(\weights) &= \sum_{i=1}^d \theta_i^2 = \| \weights \|_2^2.
\end{align}

\paragraph{Elastic net}
A linear combination of the $\ell_1$ and $\ell_2$ penalties has been also introduced by \newcite{hastieElasticNet}, called \emph{elastic net}.
Although $\ell_1$ and elastic net can be very effective in terms of sparsity, the accuracy achieved by these regularizers can be low.
On the contrary, $\ell_2$ can deliver sufficient accuracy at the cost of zero sparsity.
The need for new methods that outperform the aforementioned approaches in both accuracy and sparsity is evident.

\paragraph{Group structured regularization}
In many problems a predefined grouping structure exists within the explanatory variables, and it is natural to incorporate the prior knowledge so that the support of the model should be a union over some subset of these variable groups. 
Group structured regularization has been proposed to address the problem of overfitting, given we are provided with groups of features.
Group lasso is a special case of group regularization proposed by \citet{yuan_model_2006}, to avoid large $\ell_2$ norms for groups of weights, given we are provided with groups of features.
The main idea is to penalize together features that may share some properties.

Group structured regularization or variable group selection problem is a well-studied problem, based on minimizing a loss function penalized by a regularization term designed to encourage sparsity at the variable group level. 
Specifically, a number of variants of the $\ell_1$-regularized lasso algorithm \cite{tibshirani_regression_1996} have been proposed for the variable group selection problem, and their properties have been extensively studied recently. 
First, for linear regression, \citet{yuan_model_2006} proposed the group lasso algorithm as an extension of lasso, which minimizes the squared error penalized by the sum of $\ell_2$-norms of the group variable coefficients across groups.
Here the use of $\ell_2$-norm within the groups and $\ell_1$-norm across the groups encourages sparsity at the group level. 

In addition, group lasso has been extended to logistic regression for binary classification, by replacing the squared error with the logistic error \cite{kim2006blockwise, meier2008group}, and several extensions thereof have been proposed \cite{roth2008group}.

Later, sparse group lasso and overlapping group lasso were introduced \cite{obozinski2011group} to additionally penalize features inside the groups, while the latter can be used when groups include features that can be shared between them.

\begin{figure}[t]
\centering
\resizebox{\linewidth}{!} {

\begin{tikzpicture}[-{Stealth[length=2pt]}]

\begin{scope} [start chain, node distance=-.5pt]
\node[MyStyle, on chain] (vlos1) {};
\node[MyStyle, on chain] (vlos2) {};
\node[MyStyle, on chain] (vlos3) {};

\node[MyStyle, on chain,fill=gray] (vlos4) {};
\draw[thick,line width=2mm] (vlos4.south west) |- (vlos4.north west);
\node[MyStyle, on chain,fill=gray] (vlos5) {};
\node[MyStyle, on chain,fill=gray] (vlos6) {};
\node[MyStyle, on chain,fill=gray] (vlos7) {};
\node[MyStyle, on chain,fill=gray] (vlos8) {};
\node[MyStyle, on chain,fill=gray] (vlos9){};
\draw[thick,line width=2mm] (vlos9.south west) |- (vlos9.north west);
\node[MyStyle, on chain,fill=gray] (vlos10) {};
\node[MyStyle, on chain,fill=gray] (vlos11) {};
\node[MyStyle, on chain,fill=gray] (vlos12) {};
\node[on chain,scale=3] (vlos13) {$\cdots$};
\node[MyStyle, on chain] (vlos14) {};
\node[MyStyle, on chain] (vlos15) {};
\node[MyStyle, on chain] (vlos16) {};
\node[scale=3, on chain,font=\large] (vlos17) {\large group lasso};
\end{scope}
;

\draw [decorate,decoration={brace,amplitude=20pt,aspect=0.5}]
(vlos1.north west) -- (vlos3.north east) node[black,midway,above=8pt,scale=3]
{$G_{1}$};

\draw [decorate,decoration={brace,amplitude=20pt}]
(vlos3.north east) -- (vlos8.north east) node[black,midway,above=8pt,scale=3]
{$G_{2}$};

\draw [decorate,decoration={brace,amplitude=20pt}]
(vlos8.north east) -- (vlos12.north east) node[black,midway,above=8pt,scale=3]
{$G_{3}$};

\draw [decorate,decoration={brace,amplitude=20pt}]
(vlos14.north west) -- (vlos16.north east) node[black,midway,above=8pt,scale=3]
{$G_{L}$};

\end{tikzpicture}

}
\noindent\makebox[\linewidth]{\rule{\linewidth}{0.4pt}}

\resizebox{\linewidth}{!} {

\begin{tikzpicture}[-{Stealth[length=2pt]}]
\begin{scope} [start chain, node distance=-.5pt]
\node[MyStyle, on chain] (vlos1) {};
\node[MyStyle, on chain] (vlos2) {};
\node[MyStyle, on chain] (vlos3) {};
\node[MyStyle, on chain] (vlos4) {};
\draw[thick,line width=2mm] (vlos4.south west) |- (vlos4.north west);
\node[MyStyle, on chain,fill=gray] (vlos5) {};
\node[MyStyle, on chain,fill=gray] (vlos6) {};
\node[MyStyle, on chain] (vlos7) {};
\node[MyStyle, on chain,fill=gray] (vlos8) {};
\node[MyStyle, on chain,fill=gray] (vlos9){};
\draw[thick,line width=2mm] (vlos9.south west) |- (vlos9.north west);
\node[MyStyle, on chain] (vlos10) {};
\node[MyStyle, on chain,fill=gray] (vlos11) {};
\node[MyStyle, on chain] (vlos12) {};
\node[on chain,scale=3] (vlos13) {$\cdots$};
\node[MyStyle, on chain] (vlos14) {};
\node[MyStyle, on chain] (vlos15) {};
\node[MyStyle, on chain] (vlos16) {};
\node[minimum width=2cm,scale=2.8,on chain,align=center,font=\large,yshift=0.15cm] (vlos17) {sparse group\\lasso};
\end{scope}
;

\draw [decorate,decoration={brace,amplitude=20pt,aspect=0.5}]
(vlos1.north west) -- (vlos3.north east) node[black,midway,above=8pt,scale=3]
{$G_{1}$};

\draw [decorate,decoration={brace,amplitude=20pt}]
(vlos3.north east) -- (vlos8.north east) node[black,midway,above=8pt,scale=3]
{$G_{2}$};

\draw [decorate,decoration={brace,amplitude=20pt}]
(vlos8.north east) -- (vlos12.north east) node[black,midway,above=8pt,scale=3]
{$G_{3}$};

\draw [decorate,decoration={brace,amplitude=20pt}]
(vlos14.north west) -- (vlos16.north east) node[black,midway,above=8pt,scale=3]
{$G_{L}$};

\end{tikzpicture}
}
\noindent\makebox[\linewidth]{\rule{\linewidth}{0.4pt}}

\resizebox{\linewidth}{!} {

\begin{tikzpicture}[
  -{Stealth[length=2pt]}
  ]
   \begin{scope} [start chain, node distance=-.5pt]
    \node[MyStyle, on chain] (vlos1) {};
      \node[MyStyle, on chain] (vlos2) {};
      \node[MyStyle, on chain] (vlos3) {};
      \node[MyStyle, on chain,fill=gray] (vlos4) {};
      \node[MyStyle, on chain,fill=gray] (vlos5) {};
      \node[MyStyle, on chain,fill=gray] (vlos6) {};
      \node[MyStyle, on chain,fill=gray] (vlos7) {};
      \node[MyStyle, on chain,fill=gray] (vlos8) {};
      \node[MyStyle, on chain,fill=gray] (vlos9){};
      \node[MyStyle, on chain,fill=gray] (vlos10) {};
      \node[MyStyle, on chain,fill=gray] (vlos11) {};
      \node[MyStyle, on chain,fill=gray] (vlos12) {};
      \node[on chain,scale=3] (vlos13) {$\cdots$};
      \node[MyStyle, on chain] (vlos14) {};
      \node[MyStyle, on chain] (vlos15) {};
      \node[MyStyle, on chain] (vlos16) {};
      \node[minimum width=2.3cm,scale=2.8,on chain,align=center,font=\large,yshift=0.2cm] (vlos17) {overlapping\\group lasso};
  \end{scope}
  ;

 \draw [decorate,decoration={brace,amplitude=40pt,aspect=0.5}]
  (vlos1.north west) -- (vlos4.north east) node[black,midway,above=30pt,scale=3]
  {$G_{1}$};

  \draw [decorate,decoration={brace,amplitude=40pt}]
  (vlos3.north east) -- (vlos8.north east) node[black,midway,above=30pt,scale=3]
  {$G_{2}$};

  \draw [decorate,decoration={brace,amplitude=40pt}]
  (vlos7.north east) -- (vlos12.north east) node[black,midway,above=30pt,scale=3]
  {$G_{3}$};

   \draw [decorate,decoration={brace,amplitude=30pt}]
  (vlos14.north west) -- (vlos16.north east) node[black,midway,above=30pt,scale=3]
  {$G_{L}$};

\end{tikzpicture}

}
\noindent\makebox[\linewidth]{\rule{\linewidth}{0.4pt}}

\resizebox{\linewidth}{!} {
  

\begin{tikzpicture}[
  -{Stealth[length=2pt]}
  ]
   \begin{scope} [start chain, node distance=-.5pt]
    \node[MyStyle, on chain] (vlos1) {};
      \node[MyStyle, on chain] (vlos2) {};
      \node[MyStyle, on chain] (vlos3) {};
      \node[MyStyle, on chain] (vlos4) {};
      \node[MyStyle, on chain,fill=gray] (vlos5) {};
      \node[MyStyle, on chain,fill=gray] (vlos6) {};
      \node[MyStyle, on chain] (vlos7) {};
      \node[MyStyle, on chain,fill=gray] (vlos8) {};
      \node[MyStyle, on chain,fill=gray] (vlos9){};
      \node[MyStyle, on chain] (vlos10) {};
      \node[MyStyle, on chain,fill=gray] (vlos11) {};
      \node[MyStyle, on chain] (vlos12) {};
      \node[on chain,scale=3] (vlos13) {$\cdots$};
      \node[MyStyle, on chain] (vlos14) {};
      \node[MyStyle, on chain] (vlos15) {};
      \node[MyStyle, on chain] (vlos16) {};
      \node[minimum width=2.3cm,scale=2.8,on chain,align=center,font=\large,yshift=0.4cm] (vlos17) {overlapping\\sparse\\group lasso};
  \end{scope}
  ;

 \draw [decorate,decoration={brace,amplitude=40pt,aspect=0.5}]
  (vlos1.north west) -- (vlos4.north east) node[black,midway,above=30pt,scale=3]
  {$G_{1}$};

  \draw [decorate,decoration={brace,amplitude=40pt}]
  (vlos3.north east) -- (vlos8.north east) node[black,midway,above=30pt,scale=3]
  {$G_{2}$};

  \draw [decorate,decoration={brace,amplitude=40pt}]
  (vlos7.north east) -- (vlos12.north east) node[black,midway,above=30pt,scale=3]
  {$G_{3}$};

   \draw [decorate,decoration={brace,amplitude=30pt}]
  (vlos14.north west) -- (vlos16.north east) node[black,midway,above=30pt,scale=3]
  {$G_{L}$};

\end{tikzpicture}

}
\vspace{-0.5cm}
\caption{A graphical representation of different group lasso architectures. Grey boxes depict active features. 
\label{group-lasso}}

\end{figure}
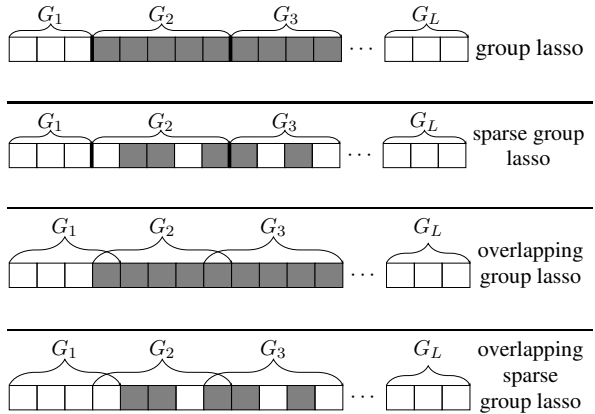

In Figure \ref{group-lasso}, we illustrate the selection of features by the most used group lasso regularizers.
In group lasso, a group of features is selected and all its features are used.
Next, in the sparse group lasso case,  groups of features are selected again but not all the features belonging to them are used. 
In the overlapping group lasso, groups can share features between them.
Finally, we may have sparse group lasso with overlaps.

\paragraph{Linguistic structured regularizers}

As mentioned previously, words that appear together in the same context, share topics or even have a similar meaning, may form groups that capture semantic or syntactic prior information.
Hence we can feed these groups to group lasso.
\newcite{yogatama_linguistic_2014} used the Alternating Direction Method of Multipliers algorithm (ADMM) \cite{boyd2011distributed} for group lasso, an algorithm that solves convex optimization problems by breaking them into smaller pieces.
In this work, groups extracted by Latent Dirichlet Allocation (LDA) and sentences were used for structured regularization.
Next, \citet{skianis2016regularizing} extended their work by utilizing topics extracted by Latent Semantic Indexing (LSI) \cite{deerwester_indexing_1990}, communities in Graph-of-Word (GoW) \cite{Rousseau:2013} and clusters in the word2vec (w2v) \cite{mikolov_distributed_2013} space.
They also performed a computational analysis in terms of the number and size of groups and how it can affect learning times.

While current state-of-the-art methods either focus on finding the most meaningful groups of features or how to further ``optimize" the group lasso approach, the attempts carry as well the disadvantages of group lasso architectures.
In some cases, we may not be able to extract ``good" groups of words.
As presented in the next section, we want to explore new ways of regularization on groups, diverging from group lasso, that can give high accuracy with high sparsity.

\section{OMP for Text Classification}
\label{sec:OMP}

The vanilla Matching Pursuit (MP) algorithm \cite{mallat1993matching} has its origin in signal processing  where it is mainly used in the compressed sensing task.
Actually, it approximates the original ``signal'' iteratively improving the current solution by minimizing the norm of the residual (approximation error). 
It can also be considered as a forward greedy algorithm for feature selection (dictionary learning problem), that at each iteration uses the correlation between the residual and the candidate features to (greedily) decide which feature to add next. The correlation between the residual and the candidate features is considered to be the length of the orthogonal projection.
Then, it subtracts off the correlated part from the residual and performs the same procedure on the updated residual.
The algorithm terminates when the residual is lower than a predefined threshold.
The final solution is obtained by combining the selected features weighted by their respective correlation values, which are calculated at each iteration.

Orthogonal Matching Pursuit \cite{Pati93orthogonalmatching} is one of the most famous extensions of the matching pursuit algorithm.
Similar to MP, OMP can be used for the dictionary learning task where it constitutes a competitive alternative to lasso algorithm.
The way it differs from the standard MP is that at every step, all the coefficients extracted so far are updated, by computing the orthogonal projection of the data onto the set of features selected so far.
In this way, the newly derived residual is orthogonal to not only the immediately selected feature at the current iteration, but also to all the features that have already been selected.
Therefore, OMP never selects the same feature twice. \citet{Tropp2004} provided a theoretical analysis of OMP, which has been generalized by \citet{Zhang09} on the stochastic noise case.

\begin{algorithm}[t]
  \caption{Logistic-OMP \label{alg:omp}}
  \small{
    \begin{algorithmic}[1]
      \REQUIRE $\MX = [\vx_1, ...,\vx_N]^{\top} \in \Reals^{N \times d}$, $\vy \in \{-1, 1\}^{N}$, $K$ (budget), $\epsilon$ (precision), $\lambda$ (regularization factor).\\
      \ENSURE $\Set = \emptyset$, $\vr^{(0)} = \vy$, $k=1$\;
      \WHILE{ $|\Set| \leq K$}
      \STATE $j^{(k)} = \argmax_{j \notin \Set} \left| \MX_j^{\top} \vr^{(k-1)}\right|$
      \IF{$| \MX_{j^{(k)}}^{\top} \vr^{(k-1)}| \leq \epsilon$}
      \BREAK
      \ENDIF
      \STATE $\Set = \Set \cup \{j^{(k)}\}$
      \STATE $\weights^{(k)} = \argmin_{\weights}{\sum_{i=1}^{N}{\mathcal{L}(\example_i, \weights, y_i)}} + \lambda\| \weights \|_2^2$  $\qquad s.t. \quad supp(\weights) \subseteq \Set$
      \STATE $\vr^{(k)} =  \frac{1}{1 + \exp\{-\MX \weights^{(k)}\}} - \mathbbm{1}_{\{\vy\}}$
      \STATE $k \pluseq 1$
      \ENDWHILE
      \RETURN $\weights^{(k)}, \Set$
    \end{algorithmic}
  }
\end{algorithm}

In the following part, we explain the main steps of the logistic OMP algorithm in detail.
Given a training set, we define $\MX = [\vx_1, ...,\vx_N]^{\top}  \in \Reals^{N \times d}$ to be the (dictionary) matrix of features (or variables) vectors, with each column $\MX_j$ to represent a feature, $\vf_j \in \Reals^N$. Let also $\vy = [y_1,\dots,y_N]^{\top}$ denote the response vector. For any set of indices $\Set$, let $\MX_{\Set}$ denote a subset of features from $\MX$, such that feature $\vf_j$ is included in $\MX_{\Set}$ if $j \in \Set$. Thus, $\MX_{\Set} = \{ \vf_j , j \in \Set \}$, with the columns $\vf_j$ to be arranged in ascending order.

OMP starts by setting the residual equal to the response vector, $\vr^{(0)} = \vy$, assuming that the set of indices $\Set$ (contains the indices of the active features) is initially empty. At each iteration $k$, OMP activates the feature that has the maximum correlation with the residual $\vr^{(k-1)}$ (calculated in the previous step):
\begin{equation}
  j^{(k)} = \argmax_{j \notin \Set} \left| \MX_j^{\top} \vr^{(k-1)}\right|.
\end{equation}
Then, we incorporate the index $j^{(k)}$ to the set $\Set$, \ie $\Set = \Set \cup \{j^{(k)}\}$. Afterwards, we apply the ordinary logistic regression by considering only the \emph{active} features. More specifically, we get the optimal coefficients by minimizing the negative log likelihood along with an $\ell_2$ penalty term:
\begin{align}
  \weights^{(k)} = \argmin_{\weights}{\sum_{i=1}^{N}{\mathcal{L}(\example_i, \weights, y_i)}} + \lambda\|\weights\|_2^2,& \nonumber \\
  \quad s.t. \quad supp(\weights) \subseteq \Set&
\end{align}
where $supp(\weights) = \{j : \theta_j \neq 0\}$. Roughly speaking, the values of the coefficients correspond to inactive features (indices) forced to be equal to zero.
Finally, we calculate the updated residual:
\begin{equation}
  \vr^{(k)} =  \frac{1}{1 + \exp\{-\MX \weights^{(k)}\}} - \mathbbm{1}_{\{\vy\}},
\end{equation}
where $\mathbbm{1}_{\{\vy\}} \defn \mathbbm{1}\{y_i \in \{1\}, \forall i \in \{1,\dots,n\}\}$ indicates if instance $\vx_i$ belongs to class $1$ or not.
We repeat the process until the residual becomes smaller than a predefined threshold, $\epsilon \geq 0$, or a desired number of active features, $K$ (budget), has been selected.
Through our empirical analysis we set $\epsilon = 0$, examining only the number of active features.
An overview of logistic-OMP is given in Alg.~\ref{alg:omp}. A detailed analysis of the algorithm's complexity is provided by \citet{Tropp:2007}.

\subsection{Overlapping Group OMP}

The Group OMP (GOMP) algorithm was originally introduced by \citet{swirszcz_grouped_2009} for linear regression models, and extended by \citet{LozanoSA11} in order to select groups of variables in logistic regression models.
Following the notion of group lasso, GOMP utilizes prior knowledge about groups of features in order to penalize large weights in a collective way. 
Given that we have words sharing some properties, we can leverage these grouping for regularization purposes.


\begin{algorithm}[t]
  \caption{Logistic Overlapping GOMP\label{alg:gomp}}
  \small{
    \begin{algorithmic}[1]
      \REQUIRE $\MX = [\vx_1, ...,\vx_N]^{\top} \in \Reals^{N \times d}$, $\vy \in \{-1, 1\}^{N}$, $\{G_1,\dots,G_J\}$ (group structure), $K$ (budget), $\epsilon$ (precision), $\lambda$ (regularization factor).
      \ENSURE $\Set = \emptyset$, $\vr^{(0)} = \vy$, $k=1$\;
      \WHILE{$|\Set| \leq K$}
      \STATE $j^{(k)} = \argmax_{j} \frac{1}{|G_j|}\left\| \MX_{G_j}^{\top} \vr^{(k-1)} \right\|_2^2$
      \IF{$\left\| \MX_{G_{j^{(k)}}}^{\top} \vr^{(k-1)} \right\|_2^2 \leq \epsilon$}
      \BREAK
      \ENDIF
      \STATE $\Set = \Set \cup \{G_{j^{(k)}}\}$
      \FOR{$i = 1$ \TO $J$}
      \STATE $G_i = G_i \setminus G_{j^{(k)}}$
      \ENDFOR
      \STATE $\weights^{(k)} = \argmin_{\weights}{\sum_{i=1}^{N}{\mathcal{L}(\example_i, \weights, y_i)}} + \lambda\| \weights \|_2^2$  $\qquad s.t. \quad supp(\weights) \subseteq \Set$
      \STATE $\vr^{(k)} =  \frac{1}{1 + \exp\{-\MX \weights^{(k)}\}} - \mathbbm{1}_{\{\vy\}}$
      \STATE $k \pluseq 1$ 
      \ENDWHILE
      \RETURN $\weights^{(k)}, \Set$
    \end{algorithmic}
  }
\end{algorithm}

Similar to \citet{LozanoSA11}, let us assume that a natural grouping structure exists within the variables consisting of $J$ groups $\MX_{G_1},\dots,\MX_{G_J}$, where $G_i \subset \{1,\dots,d\}$, and $\MX_{G_i} \in \Reals^{N \times |G_i|}$. The standard GOMP algorithm also assumes that the groups are disjoint,  $G_i \cap G_j = \emptyset$ for $i \neq j$. We will remove this assumption later on, by proposing the overlapping GOMP algorithm that is able to handle overlapping groups of features.
GOMP operates in the same way with OMP but instead of selecting a single feature, it selects a group of features with the maximum correlation between them and the residual:
\begin{equation}
  j^{(k)} = \argmax_{j} \left\| \MX_{G_j}^{\top} \vr^{(k-1)} \right\|_2^2.
\end{equation}
In the case where the groups are not orthonormalized (\ie $\MX_{G_j}^{\top} \MX_{G_j} = \eye_{G_j}$, where  $\eye_{G_j}$ is the identity matrix of size $\Reals^{|G_j| \times |G_j|}$), we select the best group based on the next criterion:
\begin{equation}
  \small
  j^{(k)} = \argmax_{j} \left| \left(\vr^{(k-1)}\right)^{\top}\MX_{G_j} (\MX_{G_j}^{\top} \MX_{G_j})^{-1} \MX_{G_j}^{\top} \vr^{(k-1)} \right|.
\end{equation}

During our empirical analysis, we have noticed that the aforementioned criteria benefit large groups. This becomes apparent especially in the case where the size of the groups is not balanced. In this way, groups with a large number of ``irrelevant'' features are highly probable to be added. For instance, it is more probable to add a group that consists of $2$ good features and $100$ bad features, instead of a group that contains only $2$ good features. To deal with situations like this one, we consider the average correlation between the group's features and the residual:
\begin{equation}
  j^{(k)} = \argmax_{j} \frac{1}{|G_j|}\left\| \MX_{G_j}^{\top} \vr^{(k-1)} \right\|_2^2.
\end{equation}

Overlapping GOMP extends the standard GOMP in the case where the groups of indices are overlapping, \ie $G_i \cap G_j \neq \emptyset$ for $i \neq j$. The main difference with GOMP is that each time a group becomes active, we remove its indices from each inactive group: $G_i = G_i \setminus G_{j^{(k)}}, \quad \forall i \in \{1,\dots,J\}$.
In this way, the theoretical properties of GOMP hold also in the case of the overlapping GOMP algorithm.
A sketch of the overlapping GOMP is shown in Alg.~\ref{alg:gomp}.

\section{Experiments}
\label{experiments}

\begin{table}[t]
	\small
	\centering
	\begin{tabular}{cccccc}
		\toprule
		& Dataset & Train & Dev & Test & Voc  \\
		\midrule
		\parbox[t]{2mm}{\multirow{4}{*}{\rotatebox[origin=c]{90}{20NG}}} & science & 949 & 238 & 790 & 25787\\
			& sports & 957 & 240 & 796 & 21938  \\
			& religion & 863 & 216 & 717 & 18822  \\
			& comp. & 934 & 234 & 777 & 16282 \\
		\midrule
		\parbox[t]{2mm}{\multirow{6}{*}{\rotatebox[origin=c]{90}{Sentiment}}} 
 			& vote & 1175 & 257 & 860 & 19813  \\
			& movie & 1600 & 200 & 200 & 43800  \\
			& books & 1440 & 360 & 200 & 21545 \\
			& dvd & 1440 & 360 & 200 & 21086 \\
			& electr. & 1440 & 360 & 200 & 10961 \\
			& kitch. & 1440 & 360 & 200 & 9248\\
		\bottomrule
	\end{tabular}
    \vspace{-0.2cm}
	\caption{Descriptive statistics of the datasets. \label{tbl:datasets-1}}
\end{table}

Next, we present the data, setup and results of our empirical analysis on the text classification task. 


\subsection{Datasets}

\paragraph{Topic categorization.} 
From the $20$ Newsgroups\footnote{\href{http://qwone.com/\string~jason/20Newsgroups/}{qwone.com/$\sim$jason/20Newsgroups/}} dataset, we examine four classification tasks.
We end up with binary classification problems, where we classify a document according to two related categories: i) \textit{comp.sys}: ibm.pc.hardware \vs mac.hardware; ii) \textit{rec.sport}: baseball \vs hockey; iii) \textit{sci}: med \vs space and iv) \textit{religion}: alt.atheism \vs soc.religion.christian.

\paragraph{Sentiment analysis.}
The sentiment analysis datasets we examined include movie reviews \citep{pang_sentimental_2004,ZaidanE08}\footnote{\label{note1}\href{http://www.cs.cornell.edu/\string~ainur/data.html}{cs.cornell.edu/$\sim$ainur/data.html}}, floor speeches by U.S. Congressmen deciding ``yea"/``nay" votes on the bill under discussion \citep{Thomas+Pang+Lee:06a}\textsuperscript{\ref{note1}} and product reviews from Amazon \citep{blitzer_biographies_2007}\footnote{\href{http://www.cs.jhu.edu/\string~mdredze/datasets/sentiment/}{cs.jhu.edu/$\sim$mdredze/datasets/sentiment/}}.

Table~\ref{tbl:datasets-1} summarizes statistics about the aforementioned datasets used in our experiments.
We choose small datasets intentionally, like \citet{yogatama_making_2014}, so that we can observe the regularization effect clearly.

\newcolumntype{C}[1]{>{\centering\let\newline\\\arraybackslash\hspace{0pt}}m{#1}}

\newcolumntype{K}[1]{>{\centering}p{#1}}

\begin{table*}[t]
\sisetup{
            detect-all,
            table-text-alignment = center,
            table-number-alignment = center,
            table-figures-integer = 1,
            table-figures-decimal = 3,
        }
	\centering
	\resizebox{0.95\textwidth}{!} {
	\begin{tabular}{cC{1.8cm}K{0.8cm}K{0.8cm}K{0.8cm}K{0.8cm}|K{0.9cm}|K{0.8cm}K{0.8cm}K{0.8cm}K{0.8cm}K{0.9cm}|C{1.24cm}}
		 \hline
		  & \multirow{2}{*}{Dataset} & \multirow{2}{*}{no reg} & \multirow{2}{*}{lasso} & \multirow{2}{*}{ridge} & \multirow{2}{*}{elastic} & \multirow{2}{*}{\underline{OMP}} & \multicolumn{5}{c|}{Group lasso regularizers} & \multirow{2}{*}{\underline{GOMP}} \\
		\cline{8-12}
		  & & & & & & & {LDA} & {LSI} & {sen} & GoW & {w2v} & \\
		\hline
		\parbox[t]{2mm}{\multirow{4}{*}{\rotatebox[origin=c]{90}{20NG}}} & {science} & 0.946 & 0.916 & 0.954 & 0.954 & 0.964* & \bfseries 0.968 & \bfseries 0.968* & 0.942 & 0.967* & \bfseries 0.968* & ~~0.953* \\
			& sports & 0.908 & 0.907 & 0.925 & 0.920 & 0.949* & 0.959 & 0.964* & \bfseries 0.966 & 0.959* & 0.946* & ~~0.951* \\
			& religion & 0.894 & 0.876 & 0.895 & 0.890 & 0.902* & 0.918 & 0.907* & \bfseries 0.934 & 0.911* & 0.916* & ~~0.902* \\
			& computer & 0.846 & 0.843 & 0.869 & 0.856 & 0.876* & 0.891 & 0.885* & 0.904 & 0.885* & \bfseries 0.911* & ~~0.902*\\
		\hline
		\parbox[t]{2mm}{\multirow{6}{*}{\rotatebox[origin=c]{90}{\textnormal{Sentiment}}}} 
 			& vote & 0.606 & 0.643 & 0.616 & 0.622 & 0.684* & 0.658 & 0.653 & 0.656 & 0.640 & 0.651 & ~~\bfseries{0.687*}\\
			& movie & 0.865 & 0.860 & 0.870 & 0.875 & 0.860* & \bfseries 0.900 & 0.895 & 0.895 & 0.895 & 0.890 & 0.850 \\
 			& books & 0.750 & 0.770 & 0.760 & 0.780 & 0.800 & 0.790 & 0.795 & 0.785 & 0.790 & 0.800 & \bfseries ~~0.805* \\
			& dvd & 0.765 & 0.735 & 0.770 & 0.760 & 0.785 & 0.800 & 0.805* & 0.785 & 0.795* & 0.795* & \bfseries ~~0.820* \\
			& electr. & 0.790 & 0.800 & 0.800 & 0.825 & \bfseries 0.830 & 0.800 & 0.815 & 0.805 & 0.820 & 0.815 & 0.800\\
			& kitch. & 0.760 & 0.800 & 0.775 & 0.800 & 0.825 & 0.845 & \bfseries 0.860* & 0.855 & 0.840 & 0.855* & 0.830 \\
		\hline
	\end{tabular}
	}
	\vspace{-0.25cm}
	\caption{
	Accuracy on the test sets.
	Bold font marks the best performance for a dataset, while \superscript{*} indicates statistical significance at $p < 0.05$ using micro sign test against lasso. For GOMP, we use w2v clusters and add all unigram features as individual groups.
	 \label{table:results-1}}
\end{table*}

\begin{table*}[t]
    \centering
    \scriptsize
    \vspace{0.1cm}
	\resizebox{0.95\textwidth}{!} {
	\begin{tabular}{C{0.01cm}C{0.95cm}K{0.45cm}K{0.45cm}K{0.45cm}K{0.5cm}|K{0.5cm}|K{0.4cm}cccc|c}
		 \hline
		  & \multirow{2}{*}{Dataset} & \multirow{2}{*}{no reg} & \multirow{2}{*}{lasso} & \multirow{2}{*}{ridge} & \multirow{2}{*}{elastic} & \multirow{2}{*}{\underline{OMP}}  & \multicolumn{5}{c|}{Group lasso regularizers} & \multirow{2}{*}{\underline{GOMP}} \\
		\cline{8-12}
		  & & & & & & & LDA & LSI & {sen} & GoW & w2v & \\
		\hline
		\hspace{-0.25cm} \parbox[t]{0.01cm}{\multirow{4}{*}{\rotatebox[origin=c]{90}{20NG}}} & science & 100 & \bf 1 & 100 & 63 & 2.7 & 19 & 20 & 86 & 19 & 21 & \color{blue} 5.8 \\
			& sports & 100 & \bf 1 & 100 & 5 & 1.8 & 60 & 11 & \color{blue} 6.4 & 55 & 44 & 7.7\\
			& religion & 100 & \bf 1.1 & 100 & 3 & 1.5 & 94 & 31 & 99 & 10 & 85 & \color{blue} 1.5 \\
			& computer & 100 & 1.6 & 100 & 7 & \bf 0.6 & 40 & 35 & 77 & 38 & 18 & \color{blue} 4.9 \\
		\hline
		\hspace{-0.25cm} \parbox[t]{0.01cm}{\multirow{6}{*}{\rotatebox[origin=c]{90}{Sentiment}}} 
 			& vote & 100 & \bf 0.1 & 100 & 8 & 5 & 15 & 16 & 13 & 97 & 13 & \color{blue} 1.5 \\
			& movie & 100 & 1.3 & 100 & 59 & \bf 0.9 & 72 & 81 & 55 & 90 & 62 & \color{blue} 2.3\\
 			& books & 100 & \bf 3.3 & 100 & 14 & 4.6 & 41 & 74 & 72 & 90 & 99 & \color{blue} 8.3\\
			& dvd & 100 & \bf 2 & 100 & 28 & 2.8 & 64 & \color{blue} 8 & \color{blue} 8 & 58 & 64 & 9\\
			& electr. & 100 & \bf 4 & 100 & 6 & 6.3 & 10 & \color{blue} 8 & 43 & \color{blue} 8 & 9 & 12 \\
			& kitch. & 100 & 4.5 & 100 & 79 & \bf 4.3 & 73 & 44 & 27 & 75 & 46 & \color{blue} 6.5\\
		\hline
	\end{tabular}
	}
	\vspace{-0.25cm}
	\caption{Model sizes (percentages of non-zero features in the resulting models). Bold for best, blue for best group.
	 \label{table:model-size-1}}
\end{table*}

\definecolor{amber}{rgb}{1.0, 0.75, 0.0}
\definecolor{mustard}{rgb}{1.0, 0.86, 0.35}
\definecolor{ao(english)}{rgb}{0.0, 0.5, 0.0}
\definecolor{bananayellow}{rgb}{1.0, 0.88, 0.21}
\definecolor{blue-violet}{rgb}{0.54, 0.17, 0.89}

 \begin{table}[t] 
 \centering
 \small
 \resizebox{\columnwidth}{!} {
 	\begin{tabular}{|c|c|c|}
 		\hline
 		\multirow{2}{*}{science} & lasso & \textcolor{red}{orbit}, \textcolor{ao(english)}{space}, contribute, funding, landing \\ 
 		\cline{2-3}
 		 & OMP & \textcolor{ao(english)}{space}, \textcolor{red}{orbit}, \textcolor{blue-violet}{moon}, \textcolor{amber}{planets}, \textcolor{blue}{scientifically}  \\
 		\hline
 	\end{tabular}
 }
 \vspace{-0.2cm}
 	\caption{Largest positive weights in lasso and OMP for the science subset of 20NG.
 	 \label{tbl:omp-weights}}
 \end{table}

\subsection{Experimental setup}
In our setup, as features we use unigram frequency concatenated with an additional bias term. 
We reproduce standard regularizers like lasso, ridge, elastic and state-of-the-art structured regularizers like sentence, LDA, GoW and w2v groups \cite{skianis2016regularizing} as baselines and compare them with the proposed OMP and GOMP.
We used pre-trained Google vectors introduced by \citet{mikolov_distributed_2013} and apply k-means clustering \cite{lloyd1982least} algorithm with maximum $2000$ clusters. 
For each word belonging to a cluster, we also keep the top $5$ nearest words so that we introduce overlapping groups.

For the learning part we used Matlab and specifically code provided by \citet{SchmidtFR07}.
If no pre-defined split exists, we separate the training set in a stratified manner by $80\%$ for training and $20\%$ for validation.

All the hyperparameters are tuned on the development dataset, using accuracy for evaluation.
For lasso and ridge regularization, we choose $\lambda$ from $\{10^{-2}$, $10^{-1}$, $1$, $10$, $10^{2}\}$. 
For elastic net, we perform grid search on the same set of values as ridge and lasso experiments for $\lambda_{rid}$ and $\lambda_{las}$. 
For group lasso, OMP and GOMP regularizers, we perform grid search on the same set of parameters as ridge and lasso experiments. 
In the case we get the same accuracy on the development data, the model with the highest sparsity is selected.
In GOMP we considered all individual features as separate groups of size one, along with the w2v groups.
Last but not least, in both OMP and GOMP the maximum number of features, $K$(budget), is set to $2000$.

\subsection{Results}
Table \ref{table:results-1} reports the results of our experiments on the aforementioned datasets.
The empirical results reveal the advantages of using OMP or GOMP for regularization in the text categorization task. 
The OMP regularizer performs systematically better than the baseline ones.
More specifically, OMP outperforms the lasso, ridge and elastic net regularizers in all datasets, as regards to the accuracy.
At the same time, the performance of OMP is quite close or even better to that of structured regularizers.
Actually, in the case of electronics data, the model produced by OMP is the one with the highest accuracy.
On the other hand, the proposed overlapping GOMP regularizer outperforms all the other regularizers in $3$ out of $10$ datasets.

Another important observation is how GOMP performs with different types of groups.
GOMP only requires some ``good" groups along with single features in order to achieve good accuracy. 
Smaller groups provided by LDA, LSI and w2v clusters provide a good solution and also fast computation, while others (GoW communities) can produce similar results with slower learning times.
This phenomenon can be attributed to the different structure of groups. 
While LDA and LSI have a large number of groups with small number of features in them ($1000$ groups, $10$ words per group), w2v clusters and GoW communities consist of smaller number of groups with larger number of words belonging to each group.
Nevertheless, we have reached to the conclusion that the selection of groups is not crucial for the general performance of the proposed GOMP algorithm.

\begin{table}[t]
    \centering
    \scriptsize
	\resizebox{\linewidth}{!}{
	\begin{tabular}{cccccc}
		 \toprule
		  & \multirow{2}{*}{Dataset} & CNN & FastText & Best OMP & Best\\
          & & (20eps) & (100eps) & or GOMP & Lasso \\
		\midrule
		\hspace{-0.25cm} \parbox[t]{0.01cm}{\multirow{4}{*}{\rotatebox[origin=c]{90}{20NG}}} & science & 0.935 & 0.958 & 0.964 & \bf 0.968 \\
			& sports & 0.924 & 0.935 & 0.951 & \bf 0.966 \\
			& religion & \bf 0.934 & 0.898 & 0.902 & \bf 0.934\\
			& computer & 0.885 & 0.867 & 0.902 & \bf 0.911 \\
		\midrule
		\hspace{-0.25cm} \parbox[t]{0.01cm}{\multirow{6}{*}{\rotatebox[origin=c]{90}{Sentiment}}} 
 			& vote & 0.651 & 0.643 & \bf 0.687 & 0.658\\
			& movie & 0.780 & 0.875 & 0.860 & \bf 0.900 \\
 			& books & 0.742 & 0.787 & \bf 0.805 & 0.800 \\
			& dvd & 0.732 & 0.757 & \bf 0.820 & 0.805 \\
			& electr. & 0.760 & 0.800 & \bf 0.830 & 0.820 \\
			& kitch. & 0.805 & 0.845 & 0.830 & \bf 0.860 \\
		\bottomrule
	\end{tabular}
	}
	\vspace{-0.3cm}
	\caption{Comparison in test accuracy with state-of-the-art classifiers:  CNN \cite{kim_convolutional_2014}, FastText \cite{joulin2017bag} with no pre-trained vectors. The proposed OMP and GOMP algorithms produce the highest accurate model in $4$ out of $10$ datasets.
	\label{tbl:dl-comp}}
\end{table}

Table \ref{table:model-size-1} shows the sparsity sizes of all the regularizers we tested.
As it becomes apparent, both OMP and GOMP yield super-sparse models, with good generalization capabilities.
More specifically, OMP produces sparse spaces similar to lasso, while GOMP keeps a significantly lower number of features compared to the other structured regularizers.
In group regularization, GOMP achieves both best accuracy and sparsity in two datasets (vote \& books), while group lasso only in one (sports).

In Table \ref{tbl:omp-weights} we demonstrate the ability of OMP to produce more discriminative features compared to lasso by showing the largest weights and their respective term.



\begin{figure*}[t]
    \centering
        \centering
        \includegraphics[width=\linewidth]{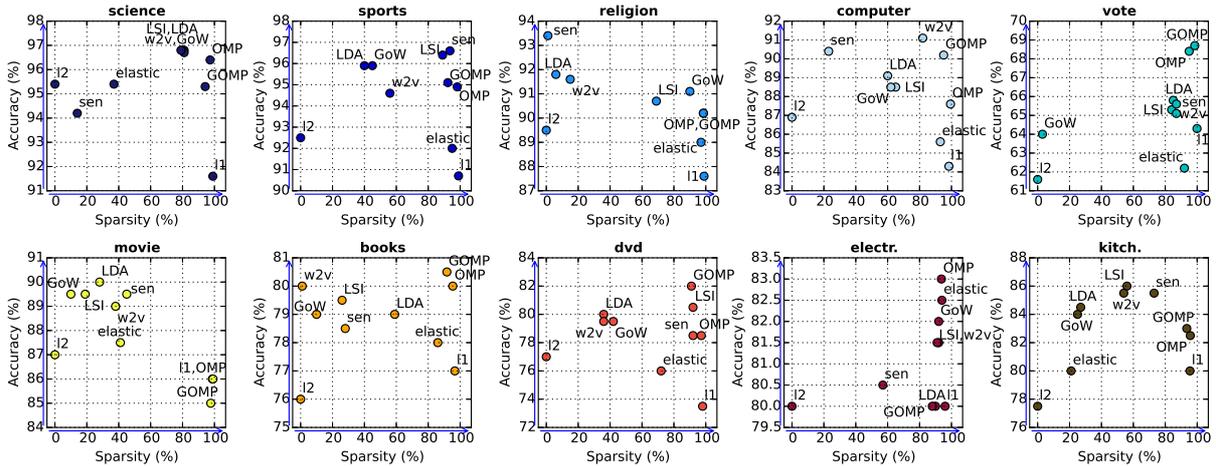}
        \vspace{-1cm} \caption{Accuracy vs sparsity on the test sets. Classifiers close to the top right corner are preferred.}
        \label{fig:acc_sparsity}
\end{figure*}

Finally, in Table \ref{tbl:dl-comp} we compare state-of-the-art group lasso classifiers with deep learning architectures \cite{kim_convolutional_2014} with Dropout \cite{srivastava2014dropout} for regularization and FastText \cite{joulin2017bag}.
We show that group lasso regularizers with simple logistic models remain very effective.
Nevertheless, adding pre-trained vectors in the deep learning techniques and performing parameter tuning would definitely increase their performance against our models, but with a significant cost in time complexity.

\subsection{Sparsity vs Accuracy}
Figure~\ref{fig:acc_sparsity} visualizes the accuracy \vs sparsity for all datasets and all classifiers.
We do that in order to identify the best models, by both metrics.
The desirable is for classifiers to belong in the top right corner, offering high accuracy and high sparsity at the same time.
We observe that OMP and GOMP tend to belong in the right parts of the plot, having very high sparsity, often comparable to the aggressive lasso, even when they do not achieve the best accuracies.

\subsection{Number of active features (atoms)}
In both OMP and GOMP algorithms, the maximum desired number of active features ($K$, budget) was used as stopping criterion.
For instance, by setting $K = 1000$, the proposed methods return the learned values that correspond to the first $\{100, 200, \dots, 1000\}$ features, respectively.
Thus, we exploit the feedforward feature selection structures of OMP and GOMP. 

\begin{figure}[t]
\includegraphics[width=\linewidth]{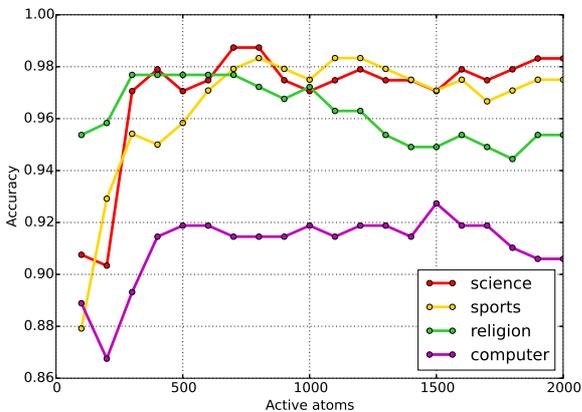} 
\vspace{-0.9cm}
\caption{Accuracy \vs number of active atoms/features for OMP on $20$NG data. \label{figure:active_atoms}}
\end{figure}

Figure~\ref{figure:active_atoms} presents the number of active features versus accuracy in the development subsets of the $20$NG dataset.
It can be easily observed that after selecting $1000$ active atoms, the accuracy stabilizes or even drops (overfitting problem). 
For instance, the best number of active features are: i) science: $700$, ii) sports: $1100$, iii) religion: $400$ and iv) computer: $1500$.
The reason for selecting $K=2000$ as the number of features to examine was to provide a sufficient number for OMP to reach a good accuracy while providing a super-sparse solution comparable to lasso.

%
%

\subsection{Time complexity}

Although certain types of group lasso regularizers perform well, they require a notable amount of time in the learning process. 



OMP offers fast learning time, given the hyperparameter values and the number of atoms.
For example, on the computer subset of the $20$NG dataset, learning models with the best hyperparameter value(s) for lasso, ridge, and elastic net took $7$, $1.4$, and $0.8$ seconds, respectively, on a $4$-core $3.00$GHz CPU.
On the other hand, OMP requires only $4$ seconds for training, making it even faster than lasso, while providing a sparser model.

GOMP can have very slow learning time when adding the features as groups individually.
This is due to the large number of groups that GOMP needs to explore in order to extract the most ``contributing" ones.
If we consider GOMP without the individual features as groups, then the learning process becomes faster, with a clear decreasing effect on accuracy.
In general, groups need to be well structured for GOMP to manage to surpass OMP and other state-of-the-art group lasso regularizers.

The advantages of the proposed methods are: (1) OMP requires no prior structural knowledge, (2) producing more discriminative features and (3) fast with relatively small number of dimensions.

Moreover, our implementation compared to the one of \citet{LozanoSA11}, provides the advantage of storing the weights and not having to recompute the whole matrices from scratch.

In the drawbacks of the methods: (1) OMP and GOMP are greedy algorithms, thus GOMP gets slow when we add the features as individual groups and (2) groups need to be ``good".

\section{Conclusion \& Future Work} 
\label{conclusion}

In this work, we introduced OMP and GOMP algorithms on the text classification task.
An extension of the standard GOMP algorithm was also proposed, which is able to handle overlapping groups.
The main advantages of both OMP and GOMP compared to other regularization schemes are their simplicity (greedy feedforward feature selection) and ability to produce accurate models with good generalization capabilities. 
We have shown that the proposed classifiers outperform standard baselines, as well as state-of-art structured regularizers in various datasets. 
Similar to \citet{NIPS2010_3974,yen2017latent,pmlr-v70-xie17b}, our empirical analysis validates that regularization remains a highly important topic, especially for deep learning models \cite{NIPS2017_6797}.

As mentioned previously, groups are not always specified in advance or hard to extract.
Especially in environments involving text.
To address this problem, we plan to extend our work by learning automatically the groups with Simultaneous Orthogonal Matching Pursuit \cite{szlam2012fast}.
Another interesting future direction would be to additionally penalize features inside the groups, similarly to sparse group lasso.
Moreover, it would be highly interesting to examine the theoretical properties of overlapping GOMP. 
Finally, as shown in recent work by \citet{NIPS2017_6797}, regularization remains an open topic for deep learning models.

\section*{Acknowledgements}
We would like to thank the anonymous reviewers for their helpful comments.

\balance

\bibliography{acl2018}
\bibliographystyle{acl_natbib_nourl}

\end{document}